\documentclass[11pt,a4paper]{article}
\usepackage[hyperref]{emnlp2018}

\usepackage{times}
\usepackage{latexsym}

\usepackage{multirow}
\usepackage{caption}
\usepackage{subcaption}
\usepackage{microtype}
\usepackage{array}
\usepackage{arydshln}
\usepackage{xcolor}
\usepackage{tabulary}
\usepackage[linesnumbered,ruled,vlined]{algorithm2e}
\usepackage{url}
\usepackage{fixltx2e}
\usepackage{epstopdf}
\usepackage{amssymb}
\usepackage{amsmath}
\usepackage{xspace,relsize,arydshln,gensymb}
\usepackage{multirow}
\usepackage{graphics,epsfig,boxedminipage}
\usepackage{adjustbox}
\PassOptionsToPackage{hyphens}{url}
\usepackage{hyperref}
\usepackage{paralist}

\makeatletter
\g@addto@macro{\UrlBreaks}{\UrlOrds}
\makeatother

\usepackage{latexsym}
\usepackage{graphicx}
\usepackage{fixltx2e}
\usepackage{epstopdf}
\usepackage{amssymb}
\usepackage{amsmath}
\usepackage{xspace,relsize,gensymb}
\usepackage{graphics,boxedminipage}
\usepackage{adjustbox,paralist}
\usepackage{booktabs}

\usepackage{soul}
\usepackage{color}
\usepackage{adjustbox,booktabs,enumitem}
\usepackage[utf8]{inputenc}
\usepackage{pgfplots}
\usepackage{array}
\newcommand{\dataset}[1]{\textsc{#1}\xspace}

\newcommand{\class}[1]{\texttt{#1}\xspace}
\newcommand{\metric}[1]{\textsf{\small #1}\xspace}

\newcommand{\method}[2][]{\texttt{#2}$_{\text{#1}}$\xspace}

\newcommand{\bert}{\method{BERT}}

\newcommand{\fev}{\method{FEVER}}
\newcommand{\eks}{KS}
\newcommand{\clf}{\method{CLIMATE-FEVER}}
\newcommand{\dpr}{\method{DPR}}
\newcommand{\bm}{\method{BM25}}
\newcommand{\fid}{\method{FiD}}
\newcommand{\bpr}{\method{BPR}}
\newcommand{\tfive}{\method{T5}}
\newcommand{\tfivebaseline}{\method{T5-only}}
\newcommand{\topone}{\method{Top1}}
\newcommand{\clex}{\method{Bert-veracity}}

\newcommand{\brscore}{\metric{B-SCORE$_{\text{\sf rs}}$}}
\newcommand{\rgone}{\metric{ROUGE-1}}
\newcommand{\rgl}{\metric{ROUGE-L}}
\newcommand{\vag}{\metric{V-AGREE}}
\newcommand{\mar}{\metric{MAR}}
\newcommand{\acc}{\metric{ACC}}
\newcommand{\fevthree}{\method{FEV3}}
\newcommand{\fevtwo}{\method{FEV2}}
\newcommand{\clfev}{\dataset{C-fever}}
\newcommand{\clfeed}{\dataset{Feedback}}
\newcommand{\anone}{\method{Anno-T1}}
\newcommand{\antwo}{\method{Anno-T2}}
\newcommand{\wiki}{\dataset{Wiki}}
\newcommand{\papers}{\dataset{Pubs}}

\newcommand{\ftof}{\clfev{}$\rightarrow$\clfev}
\newcommand{\ftofeed}{\clfev{}$\rightarrow$\clfeed}
\newcommand{\feedtof}{\clfeed{}$\rightarrow$\clfev}

\newcommand{\ex}[1]{\textit{#1}\xspace}

\newcommand{\secref}[2][]{Section#1~\ref{sec:#2}}
\newcommand{\figref}[2][]{Figure#1~\ref{fig:#2}}
\newcommand{\tabref}[2][]{Table#1~\ref{tab:#2}}

\newcommand\email{\begingroup \urlstyle{tt}\smaller\Url}

\aclfinalcopy 

\title{Automatic Claim Review for Climate Science via Explanation Generation}

\author{Shraey Bhatia \qquad Jey Han Lau \qquad Timothy Baldwin \\[1ex]
    School of Computing and Information Systems,\\The University 
   of
Melbourne \\[0.9ex]
  \email{shraeybhatia@gmail.com}, \email{jeyhan.lau@gmail.com}, 
 \email{tb@ldwin.net}}
  \date{}

\begin{document}

\maketitle

\begin{abstract}
There is unison is the scientific community about human induced climate change. Despite this, we see the web awash with claims around climate change scepticism, thus driving the need for fact checking them but at the same time providing an explanation and justification for the fact check. Scientists and experts have been trying to address it by providing manually written feedback for these claims. In this paper, we try to aid them by automating generating explanation for a predicted veracity label for a claim by deploying the approach used in open domain question answering of a fusion in decoder augmented with retrieved supporting passages from an external knowledge. We experiment with different knowledge sources, retrievers, retriever depths and demonstrate that even a small number of high quality manually written explanations can help us in generating good explanations.
\end{abstract}

\section{Introduction}
\label{sec:intro}

\begin{figure}[t]
\centering
\includegraphics[width=\linewidth]{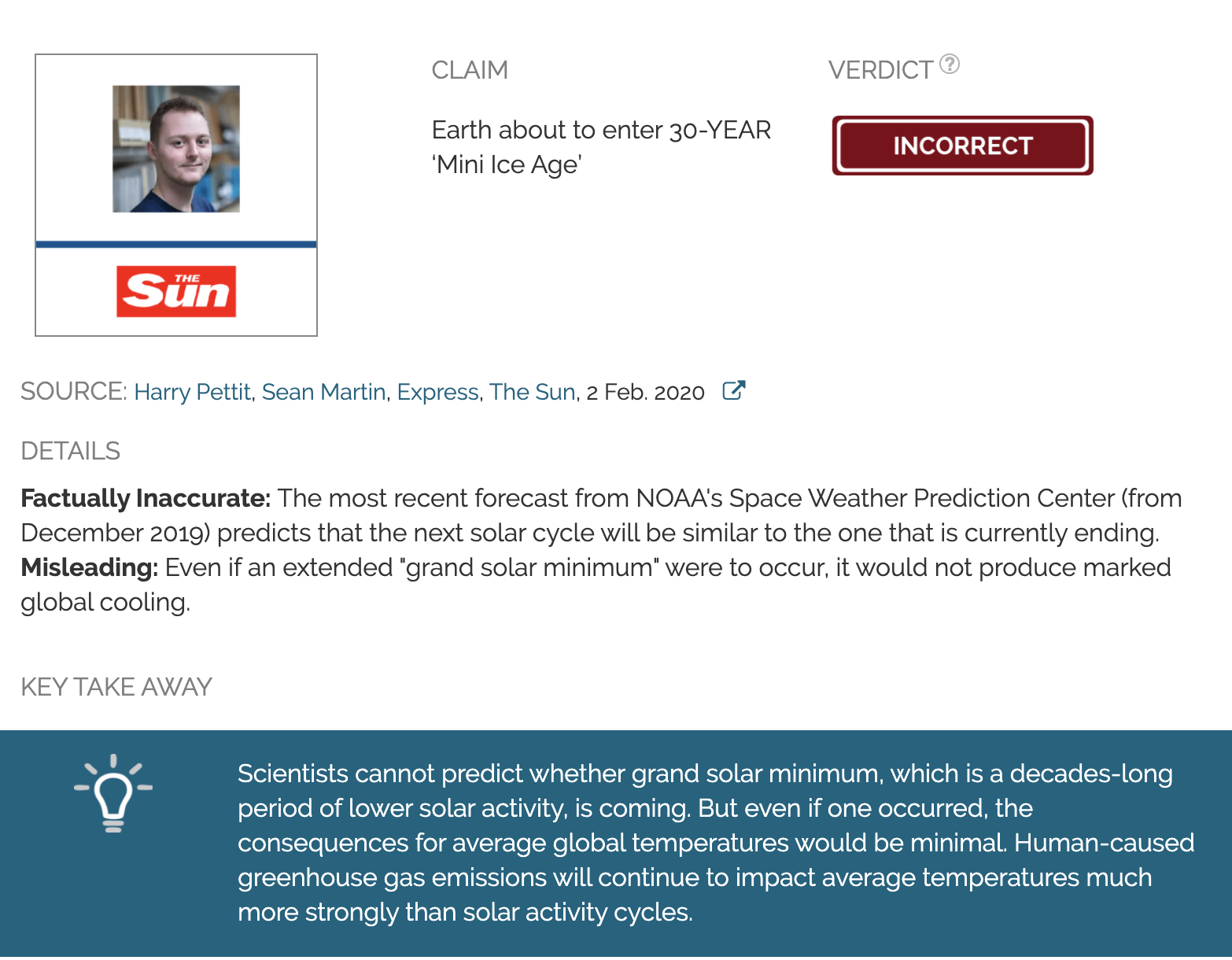}
\caption{An example of a claim review from \url{climatefeedback.org}}
\label{fig:cf-ex}
\end{figure}

Climate change is one of the biggest challenges threatening the world,
the effects of which are already being felt through events such as
increasingly frequent extreme weather effects, severe droughts, and
devastating fires in countries such as the US and Australia
\cite{nhess+:2020}.  During such events, it is not uncommon to see
claims of questionable scientific merit with headlines such as
\ex{Climate Change has caused more rain, helping fight Australian
  wildfires}.\footnote{\url{https://bit.ly/2H9xivN}} This kind of
narrative seeds scepticism \cite{oreskes+:2010}, discredits climate
science and scientists \cite{anderegg+:2010}, spreads misinformation
\cite{farrell+:2019}, and neutralises debate on key issues
\cite{mckie+:2018}, thereby turning it into a partisan issue
\cite{benegal+:2018, van+:2017} and leading to inaction. To check for
the veracity (truthfulness) of such claims and at the same time provide
public with scientifically sound information, experts have started
publishing feedback on websites like
\href{https://climatefeedback.org/}{climatefeedback.org} and
\href{https://skepticalscience.com/}{skepticalscience.com}. \figref{cf-ex}
gives us one such example where the claim from The Sun is that \ex{Earth
  [is] about to enter 30-year `Mini Ice Age'}, which has been labelled
as Incorrect, with the Key Take Away being that \ex{Scientists cannot
  predict whether solar grand minimum ... is coming} and \ex{even if one
  occurred, the consequences for average global temperatures would be
  minimal}.  It is this process of fact verification with a textual
explanation/justification that we aim to automate, as a tool to assist
climate science experts to more efficiently respond to such claims.

Our approach draws on recent work on explainable fact checking
\cite{atanasova+:2020} and retrieval-augmented generation
\cite{lewis+:2020}, in using the claim to: (1) retrieve documents from a
knowledge source such as Wikipedia or Intergovernmental Panel on Climate
Change (IPCC) reports; and (2) generate an explanation for the claim
based on the top-$k$ retrieved documents and the T5 decoder
\cite{raffel+:2019}, with a multi-task objective including a veracity
prediction for the claim.

Our contributions are as follows: (1) we introduce the task of
generating explanations justifying the predicted veracity label for a
climate change claim; (2) we deploy open-domain question answering with
an external knowledge source to the knowledge-rich and high-impact
domain of climate change fact checking; (3) we study the effect of
different knowledge sources, retrievers, and retrieval depth across two
datasets (both within- and cross-dataset); and (4) we demonstrate that small
numbers of manually-written high-quality claim explanations result in
high-quality explanations.

\section{Related Work}
\label{sec:related-work}

Public perception and reaction to climate change is a function of how
the facts and narrative are presented \cite{flottum+:2014,
  flottum+:2016}, in large part because climate change is not just the
science but has strong political, social, and ethical aspects
\cite{flottum+:2017}.  A range of corpus linguistic methods have been
used to study the topical and stylistic aspects of language around
climate change, including structured topic modelling
\cite{roberts+:2014,tvinnereim+:2015}, keyphrase extraction via grammar
induction \cite{salway+:2014}, and analysis of frequently-used metaphors
\cite{atanasova+:2017}.

Fact checking and fake news detection are critical tasks for climate
science discussions in the media and social media. Early work on fact
checking and misinformation was based on the creation of datasets for
fake news detection \cite{vlachos+:2014} and claim/stance verification
\cite{ferreira+:2016}. While early datasets were limited in size, larger
datasets have since been developed, such as LIAR \cite{wang+:2017} ---
collected from PolitiFact and labelled with 6 levels of veracity --- and
\fev \cite{thorne+:2018}, a dataset generated from Wikipedia with
\class{supported}, \class{refuted} and \class{not enough info}
labels. More recently, extending the methodology of \fev,
\newcite{diggelmann+:2020} released \clf, a dataset specific to the domain
of climate change.

Several studies have analysed the discourse around climate change. 
\newcite{luo+:2020} proposed an opinion framing task to detect stance in 
the media. To understand narrative and fames around lack of action on 
climate change, \newcite{bhatia+:2021} studied automatic classification 
of neutralization techniques. \newcite{diggelmann+:2020} formalised the 
task by introducing the task of \clf as a veracity 
prediction task in a fact checking setting. Our work differs from these in 
that we are focussed on jointly generating the correct explanation to 
counter or support the claim, in addition to the veracity label.

The closest work to our own is that on explainable fact checking by 
\newcite{atanasova+:2020}, which uses DistilBERT \cite{sanh+:2019} in 
a multitask setting and performs the joint task of
summarisation and classification of the veracity of the claim.  
\newcite{stammbach+:2020} experimented with GPT3's \cite{brown+:2020} 
few shot learning capabilities to generate fact-checking explanations.

There has recent work on the ability of pretrained models like BERT
\cite{devlin+:2019}, GPT2 \cite{radford+:2019}, and \tfive
\cite{raffel+:2019} to capture factual information
\cite{petroni+:2019}. However, ``knowledge'' in these pretrained models
is stored in the parameters and not directly accessible, making it hard
to interpret, extend, or even query these models
\cite{roberts+:2020}. One way of augmenting pretrained models is to
combine them with external knowledge sources by retrieving passages that
are similar to a given query (a claim, in our case).  The text retrieval
module can be either traditional methods such as \bm \cite{robertson+:1995}, or neural
retrievers such as dense passage retriever (\dpr) \cite{karpukhin+:2020}
or memory efficient binary passage retriever \cite{yamada+:2021}.

Retrieval-based methods have been successfully applied to open domain
question answering (QA) by combining the retriever with a ``reader'', to
extract the relevant answer from those passages.  \newcite{chen+:2017}
introduced DRQA, a span-based extractive framework trained with gold
spans in a SQuAD setting \cite{rajpurkar+:2016}, TF-IDF-weighted sparse
representations are used to retrieve relevant passages from Wikipedia,
and answers are extracted from them. \newcite{lee+:2019} argued against
using separate information retrieval systems to retrieve context
passages, and proposed ``open retrieval question answering'', which
jointly learns the reader and retriever using only QA pairs (without
explicit supervision over context passages). The retriever is pretrained in an
unsupervised setting using \textit{inverse cloze task}, i.e.\ by
predicting the document context given a passage from that document. Similarly \newcite{guu+:2020}
jointly trained a reader and retriever by pretraining
the retriever using ``salient span masking'', a specialisation of masked
language modelling.

\newcite{lewis+:2020} proposed retrieval-augmented generation, which
combines a pretrained language model with an external knowledge source
accessed via a neural retriever such as \dpr \cite{karpukhin+:2020}, and
jointly fine-tuned in a seq2seq manner (over questions and
answers). Building on a similar idea, \newcite{izacard+:2020,
  izacard+:2020b} proposed the simple but highly effective ``fusion-in
decoder'' model, which combines evidence from multiple passages
independently in separate encoders, and attending to the combined
representations in the decoder to generate the
answer. \newcite{samarinas+:2021} extended the idea of passage retrieval
to automatic fact checking, and demonstrated that neural retrieval
models can improve evidence recall. In our work, we combine these ideas
to jointly perform claim veracity classification and generate
explanations to justify the prediction.

\begin{figure*}[t]
\centering
\includegraphics[width=0.85\textwidth]{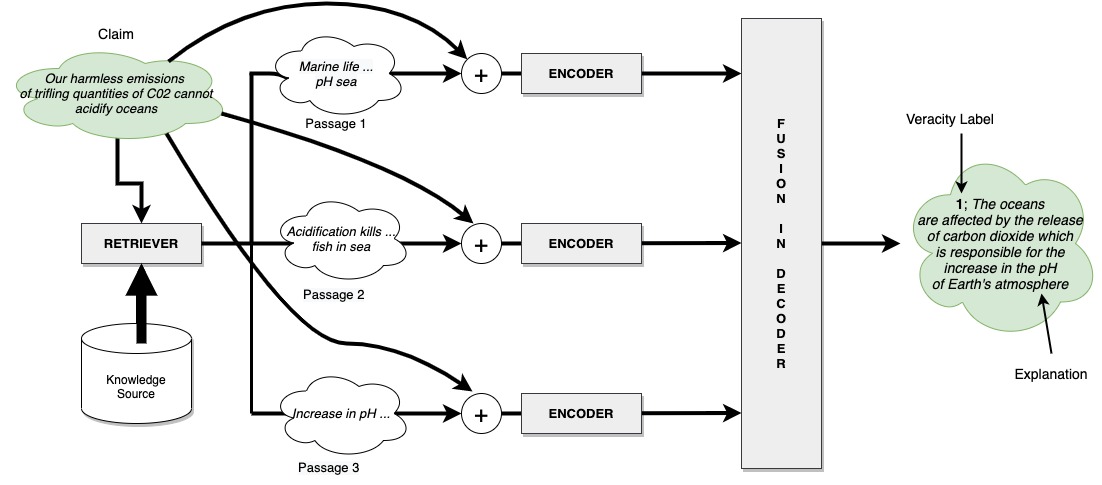}
\caption{Overview of our proposed method for generating an explanation
  and veracity label for a given claim, based on text passages from
  a knowledge source.}
\label{fig:wf}
\end{figure*}

\section{Datasets}
\label{sec:datasets}

The two key data components of our method are: (1) an external knowledge
source (``\eks''); and (2) paired claim--explanation data with
veracity labels (where the explanation justifies the binary veracity class).

\subsection{Knowledge Sources}
\label{sec:knowledge-source}

We experiment with two knowledge sources:

\paragraph{Wikipedia (``\wiki'')} Following \newcite{chen+:2017,
  karpukhin+:2020}, we use the processed Wikipedia dump from Dec 2018 as
the knowledge source.\footnote{Available via the \dpr repository:
  \url{https://github.com/facebookresearch/DPR}}

\paragraph{Peer-reviewed PubMed abstracts and IPCC reports
  (``\papers'')} A combination of climate change-related abstracts from
PubMed,\footnote{\url{https://pubmed.ncbi.nlm.nih.gov/}} and reports
from the Intergovernmental Panel on Climate Change
(IPCC).\footnote{\url{https://www.ipcc.ch/}} PubMed is a database of
peer-reviewed publications primarily in the biomedical domain, but also
including other high-profile scientific journals. We use MeSH categories
to sample publications relating to climate science, and extract the
title and abstract of each publication.
  IPCC reports are written by a mix of scientists, experts, and policy
  makers. They are based off peer-reviewed publications, and intended to
  provide a comprehensive summary of a given topic relating to topics
  such as the physical science of climate change, climate change
  impacts, or the mitigation of climate change.
  
  While smaller in size than \wiki, this knowledge source is specific to
  climate change and thus more domain relevant. 

We apply the same preprocessing steps as \newcite{chen+:2017,
  karpukhin+:2020} to both knowledge sources, and segment each document
into (non-overlapping) 100-word passages.
In total after preprocessing, we generate 21M and 123K passages for \wiki
and \papers, respectively.

\subsection{Paired Data}
\label{sec:pd}

\paragraph{\clf (``\clfev'')} \newcite{diggelmann+:2020} released this
dataset for climate change claim verification, consisting of 1535 claims
with 5 corresponding evidence sentences\footnote{``Evidence'' is used
  instead of ``explanation'' in this section for consistency with \clf,
  but both mean the same thing.} each (yielding a total of 7675
claim--evidence pairs). Following \fev \cite{thorne+:2018}, it uses
Wikipedia as the knowledge source for the evidence sentences, and labels
the veracity of each evidence sentence according to 3 classes:
\class{supports}, \class{refutes}, and \class{not enough info}. An
overall label is assigned to a claim based on a majority vote over the
evidence labels for the 5 evidence sentences.

Inspired by \newcite{thorne+:2020,lewis+:2020}, we explore 2 
configurations of \clfev in our experiments: (1) 3-way classification (``\fevthree''); and (2)
2-way classification  (``\fevtwo''), where we consider only \class{supports} 
vs.\ \class{refutes} and discard any claims which are not
majority-labelled according to one of these two labels.  For a given claim, we filter out evidence 
sentences which differ in label to the overall claim label. We 
split the two variants of \clfev into training, validation, and test sets using stratified partitioning.
This resulted in: 963 training, 83 validation and 332 test instances for
\fevthree; and 680 training, 50 validation, and 177 test instances for
\fevtwo. As each claim has multiple evidence sentences, this translates
into a total of 3196 claim--evidence pairs for \fevthree, and 1671
claim--evidence pairs for \fevtwo.  When evaluating the quality of the
generated explanation for a claim, we consider the multiple evidence
sentences as ground-truth references.

\paragraph{Climate Feedback (``\clfeed'')}
\href{https://climatefeedback.org/}{climatefeedback.org} is a website
which invites
scientists and experts
to provide reviews and highlight factual inaccuracies of claims in news
articles or made by prominent public figures. An example of a claim
review is given in \figref{cf-ex}, where the ``key-take away'' acts as
our explanation. Following \clfev, we construct a dataset of
claim--explanation pairs.  One important difference to \clfev is that
the explanations here are not passages from a document collection, but
rather are written by experts and so are more descriptive, specific to
the claim, and overall higher in quality. We crawl 130 paired
claim--explanation instances from the website. As the website is almost
exclusively used to refute incorrect claims, the vast majority of claims
have veracity label incorrect or partially incorrect. Given the
resultant extreme label imbalance, we do not use this resource for
veracity prediction, but only for explanation generation. Unlike \clfev,
there is always a unique explanation for each claim, because of the
structure of the website. In our experiments, we use five random
training/validation/test splits of the data (90/15/25), and average the results.


\section{Method}
\label{sec:method}

We now describe the joint model for explanation generation and  veracity 
prediction. Our model is based on the \textit{Fusion in Decoder} (``\fid''; 
\newcite{izacard+:2020}) --- a sequence-to-sequence model that takes as 
input a question and support passages (from a retriever), and generates an 
answer --- adapted to process claim and support passages, to predict 
veracity labels and generate explanation texts.

There are two components in our model: (1) a retriever 
(\secref{retriever}); and (2) a generator (\secref{generator}). Given a 
claim $c_i$, the role of the retriever is to search for the most 
relevant (top-$k$) support passages ${z_{k}}$ from a knowledge source 
(e.g.\  \wiki).  For the generator, it is fashioned as an encoder--decoder:
given a claim with $k$ support passages, each support passage 
${z_{k}}$ is concatenated with the claim $c_i$ to produce $k$ 
claim--passage contexts $e_{i} = [c_{i};z_{j}]$, where $1\leq j \leq k$, 
where each $e_{i}$ is encoded independently but the
encodings are fed together to the decoder to generate the veracity label 
\textit{and} explanation.  This joint processing of multiple 
claim--passage contexts in the decoder allows the model to summarise the 
evidence from multiple passages; an illustration of the overall 
architecture is presented in \figref{wf}.


\begin{table}[t]
\centering
\begin{adjustbox}{max width=\linewidth}
\begin{tabular}{lc@{\;}cc@{\;}c}
\toprule
\multirow{2}{*}{Method} & \multicolumn{2}{c}{\brscore} & 
                                  \multicolumn{2}{c}{\acc} \\ & \fevtwo               
& \fevthree               & \fevtwo                & \fevthree                 
\\ \midrule
\tfivebaseline                  & 0.26               &  0.24                   
                                                              & 0.75                 & 0.49                     \\
  \topone (\wiki)                        
& 0.03               & 0.02                   &NA                 & NA                   

  \\ \topone (\papers)                       & 0.02               & 0.03                   
                                                              &NA                 & NA                   \\
  \clex                          
& NA             & NA                   & 0.79                 & 0.55                    
\\
 \bottomrule
\end{tabular}
\end{adjustbox}
\caption{Performance of the baseline models (explanation generation =
  \brscore; veracity prediction = \acc).}
\label{tab:baseline-res}
\end{table}

\begin{table*}[t]
\centering
\begin{adjustbox}{max width=\textwidth}
\begin{tabular}{clc@{\;}cc@{\;}cc@{\;}cc@{\;}c}

\toprule

\multirow{2}{*}{{Knowledge Source}} & 
\multirow{2}{*}{{Retriever}} & \multicolumn{2}{c}{{\brscore}} & \multicolumn{2}{c}{{\rgone}} & 
\multicolumn{2}{c}{{\rgl}} & 
\multicolumn{2}{c}{{\acc}} \\

 &  & {\fevtwo} & {\fevthree} & {\fevtwo} & {\fevthree} & {\fevtwo} & {\fevthree} & {\fevtwo} & {\fevthree} \\
\midrule

\multicolumn{2}{c}{Baseline: \tfivebaseline} & 0.26 & 0.24 & 0.25 & 
0.25 & 
0.22 & 0.21 & 0.75 &0.49 \\
\multicolumn{2}{c}{Baseline: \clex}  & --- & --- & --- & --- & --- & --- 
 & 0.79 & 0.55 \\
\midrule
\multirow{2}{*}{{\wiki}}

 & \bpr              & 0.28 & 0.27 & 0.25 & 0.27 & 0.22 & 0.24 & 0.80 & 0.59 \\
 
 & \bm  & 0.26 & 0.26 & 0.26 & 0.26 & 0.23 &0.22 & 0.78 & 0.57 \\

 \midrule
 
\multirow{2}{*}{{\papers}}

 & \bpr              & 0.28 & 0.26 & 0.26 & 0.25 & 0.23 & 0.22 & 0.77 & 0.55 \\
 
 & \bm   & 0.29 & 0.26 & 0.25 & 0.25 & 0.23 &0.21 & 0.77 & 0.53 \\
 
\bottomrule

 

\end{tabular}
\end{adjustbox}
\caption{\ftof results (explanation generation =
  \brscore, \rgone, and \rgl; veracity prediction = \acc).}
\label{tab:results1}
\end{table*}

\subsection{Retriever: \bm and \bpr}
\label{sec:retriever}

We experiment with two retrievers: (1) \bm \cite{robertson+:1995}; and
(2) \bpr (binary passage retrieval; \newcite{yamada+:2021}). For \bm,
the knowledge source is stored in the form of an inverted index. Claim
texts are tokenised and entities are linked to produce a sparse bag of
words/concepts representation. We use
Pylucene\footnote{\url{https://lucene.apache.org/pylucene/}} with 
default parameters as the retrieval engine, and DBepdia
spotlight\footnote{\url{https://www.dbpedia-spotlight.org/api}} for
entity recognition and linking.

\dpr \cite{karpukhin+:2020} is a dual encoder that consists of two
separate BERT models to encode the query and passage, and computes the
relevance score based on the inner product of their BERT encodings.
\bpr extends this by integrating a hashing layer (which converts the
BERT encodings to binary codes) to making the encodings
more memory efficient without substantial loss in accuracy.  \bpr is trained with a
multi-task objective over 2 tasks: (1) candidate generation using the
binary codes; and (2) candidate re-ranking based on the inner product of
the continuous vectors.  We use the official implementation of \bpr,
which was pretrained on the Natural Questions dataset
\cite{kwiatkowski+:2019} with Wikipedia as the knowledge
source.\footnote{\url{https://github.com/studio-ousia/bpr}} 
Note that we use the pretrained \bpr without fine-tuning.


\subsection{Generator: \tfive}
\label{sec:generator}


\newcite{raffel+:2019} introduced \tfive, a pretrained encoder--decoder
model, where different NLP tasks can be reframed as text-to-text
problems to allow the training of a single model to perform multiple
tasks.  \tfive allows us to define a new task by prepending a
task-specific prefix token during fine-tuning.  In our case, an input is
prefixed with \class{lab-exp:} (to denote our task) and uses special
tokens \class{claim:} and \class{context:} to denote the start of a
claim text and support passage, respectively (e.g.\ input $=$
\textit{lab-exp: claim: Our harmless emissions of ... context: Marine
  life ...}). The output takes the form of the veracity label followed by
an explanation (delimited by a semicolon, see \figref{wf}), such that
that the decoder is predicting the veracity label and generate
explanation together.\footnote{We also experimented with defining
  separate objectives for label generation and explanation generation
  and found similar results.}

\subsection{Experimental setup}

For \fevtwo when training on \clfev, we pretrain
\tfive base as a generator as follows: batch size = 1
with gradient accumulation = 4, text maxlength (claim +passage length) =
200, and generated answer maxlength = 150. We use the Adam optimiser, learning
rate = 1e-5 with a linear scheduler, weight decay = 0.01, and total
steps = 10k with warmup steps = 800. We evaluate the performance of our
model on the validation set every 2500 steps.

In the case of \fevthree, due to the larger size of the dataset, we
change the total steps to 18k and warmup steps to 1000, but keep other
hyper-parameters the same. In the case of training on \clfeed, we
decrease our total steps to 7500. Details of \bpr and \bm are given in
\secref{retriever}. We experiment with $k \in \{1, 5, 10, 15, 20\}$ as
the number of retrieved documents for both retrievers.

\section{Experiments}


In this section we present and compare the results of our experiments 
under different conditions.  We evaluate the performance of explanation 
generation using rescaled BERT-score (\brscore; \newcite{zhang+:2019}), 
\rgone and \rgl.
The original BERT-score uses contextual embeddings to compute similarity 
between a generated explanation and reference, but since the computed 
similarity values often end up in a small range at the higher end of the 
numeric range (close to 1), \brscore is proposed where rescaling is performed to 
produce similarity scores of wider range.  To assess label veracity 
prediction, we use classification accuracy (\acc).

We use several baselines: (1) \tfivebaseline, where we remove the
retriever and treat the task as a sequence-to-sequence problem (i.e.\
\tfivebaseline is trained using only the claim-explanation pairs without
any knowledge sources); (2) \topone, where we remove the generator and
use the top-1 retrieved passage (using \bpr) as the explanation (this
baseline therefore does not predict the veracity labels); and (3) \clex,
where we fine-tune \bert using the claim$+$explanation as input to
predict the veracity label (this baseline therefore does not
have a retriever or generate any explanation).

\begin{table}[t] \centering
\begin{adjustbox}{max width=0.8\linewidth}
\begin{tabular}{lccc}
\toprule

\eks                      & \brscore & \rgone & \rgl \\
\midrule
\wiki          & 0.17                     & 0.16                  & 0.14    
\\

\papers         & 0.17                     & 0.16                  & 0.14  \\


\bottomrule
\end{tabular}
\end{adjustbox}
\caption{Generation performance for \ftofeed, with \bpr as the retriever.}
\label{tab:results2}
\end{table}

\begin{table}[t] \centering
\begin{adjustbox}{max width=0.8\linewidth}
\begin{tabular}{lccc}
\toprule

\eks                      & \brscore & \rgone & \rgl \\
\midrule
\wiki          & 0.21                     & 0.22                  & 0.18    
\\

\papers         & 0.22                     & 0.21                  & 0.18  \\


\bottomrule
\end{tabular}
\end{adjustbox}
\caption{Generation performance for \feedtof, with \bpr as the retriever.}
\label{tab:results3}
\end{table}

\begin{figure*}[t]
\begin{subfigure}{.5\textwidth}
\centering
\includegraphics[width=\textwidth]{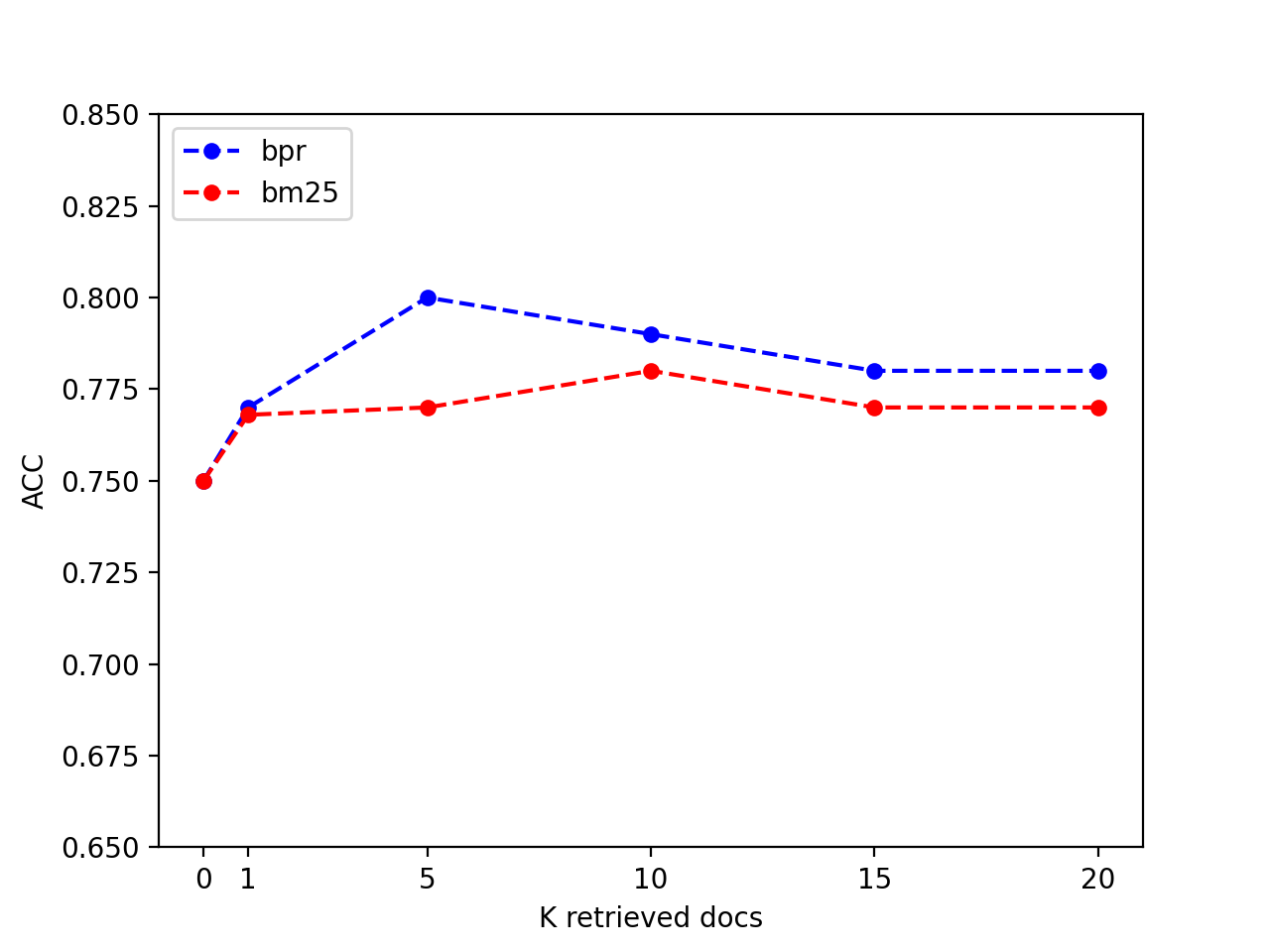}
\caption{\fevtwo}
\label{fig:res-fev2}
\end{subfigure}
~
\begin{subfigure}{.5\textwidth}
\centering
\includegraphics[width=\textwidth]{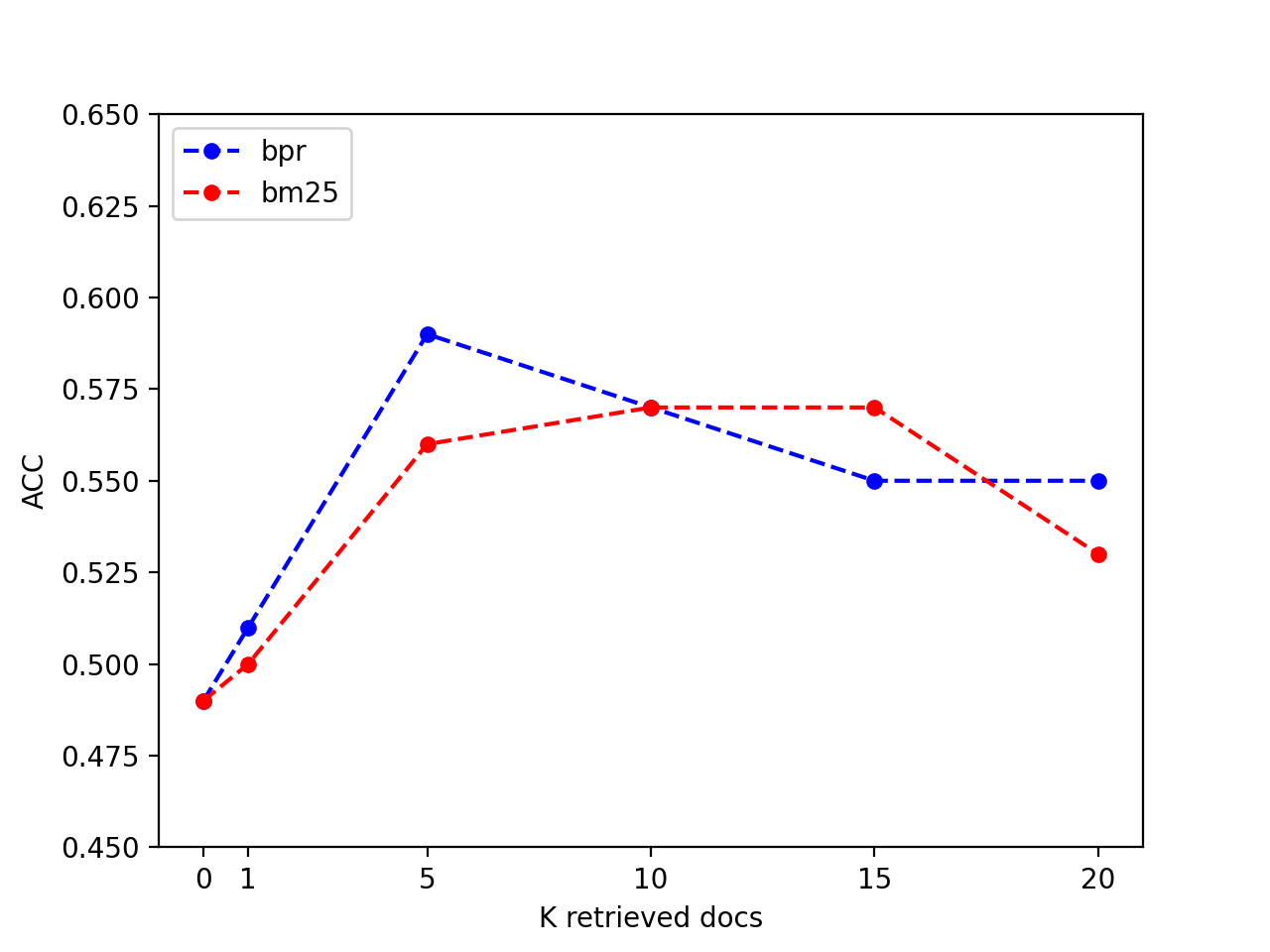}
\caption{\fevthree}
\label{fig:res-fev3}
\end{subfigure}
\caption{\acc performance over different numbers of retrieved documents for \bpr and \bm, with \wiki.}
\label{fig:res-ret}
\end{figure*}

\subsection{RQ1: Which knowledge sources performs best?}
\label{sec:rq1}

We first present the baseline system results in \tabref{baseline-res}.
Note that all results are presented in the two configurations of \clfev,
where the veracity has either 2 classes (\fevtwo) or 3 classes
(\fevthree); see \secref{pd} for details.  In terms of explanation
generation, we can see that \tfivebaseline substantially outperforms
\topone (using either \wiki or \papers). This suggests that it is
important to have a generator summarise over multiple evidence passages
to generate a good explanation. In terms of veracity prediction, we see
a pure veracity prediction model (\clex) performs best, although it is
important to note that this model uses explanation text as part of its
input (where \tfivebaseline has to generate the explanation).

\tabref{results1} presents the full results for veracity prediction and
explanation generation over \clfev (for training and testing), using
either \wiki or \papers (\secref{knowledge-source}) as the knowledge
source. In terms of veracity prediction (\acc), our model (using
either \wiki or \papers) outperforms \tfivebaseline; more encouragingly,
our best model (using \wiki and \bpr) outperforms \clex which has access to
the ground-truth explanation as input, suggesting the explanation
generator component helps veracity prediction.


In terms of explanation generation quality, our model consistently
outperforms \tfivebaseline (using either \wiki or \papers) over all
metrics (\brscore, \rgone, and \rgl), implying that the incorporation of
an external knowledge source aids explanation generation. Comparing the
results between \wiki and \papers, we see that \wiki generally performs
slightly better.  We theorise this may be due to 2 reasons: (1) \clf is
based on Wikipedia (as the source of evidence sentences), and so there
is an element of data bias when we using the \wiki as our knowledge
source;\footnote{It is theoretically possible for the retriever to
  retrieve the same supporting passage as the explanation text (output),
  given that the explanation is extracted from Wikipedia in \clfev. In
  practice this is rare since the retriever uses the claim text as the
  query.} and (2) \papers is orders of magnitude smaller in terms of the
number of passages, and its domain alignment advantage appears to be
outweighed by the resulting data sparseness.


\subsection{RQ2: Which retriever performs best?}

Looking at \tabref{results1}, in terms of both veracity prediction and 
explanation quality, \bpr generally outperforms \bm, although the 
performance gap is smaller when we use \papers as the knowledge source.
As such, we base the remainder of our experiments exclusively on \bpr.

\subsection{RQ3: How well does the method work in the cross-dataset setting?}
\label{sec:rq3}

\tabref{results1} was based on the \ftof dataset setting. Here,
we explore cross-dataset performance to test the robustness of the
proposed model, based on the two settings of: (1) train on\clfev and test on
\clfeed (\ftofeed); and (2) train on \clfeed and test on \clfev
(\feedtof).

We first present \tabref{results2}, evaluating only the generation
quality, recalling our comments about label imbalance in \clfeed from
\secref{pd}. We see a considerable drop in values in comparison to the
in-dataset setting of \tabref{results1}.
This drop can be attributed to the fact that \clfeed explanations are 
manually written, and generally longer and follow a different style 
to \clfev.

Next we look at \tabref{results3} for \feedtof and see that the raw
results are higher than \tabref{results2}, though still below the in-dataset
setting.






\begin{table*}[t]
\small
\centering
\begin{tabular*}{\linewidth}{l}\toprule
\textbf{Text} \\
\midrule
\textbf{Claim}:About 60\% of the warming observed from 1970 to 2000 was very likely caused by the above natural 60-year climatic \\ cycle during its warming phase.\\
\textbf{\wiki} The observed warming over the past 60 years was ``very likely'' (greater than 90\% probability, based on expert \\ judgement) due to human-induced emissions of greenhouse gases. \\
\textbf {Ref}: It is extremely likely (95-100\% probability) that human influence was the dominant cause of global warming between \\ 1951-2010. \\
\hdashline
\textbf{Claim}: Our harmless emissions of trifling quantities of carbon dioxide cannot possibly acidify the oceans\\
 \textbf{\papers}: The oceans are affected by the release of carbon dioxide, is responsible for the increase in the pH of the \\ Earth's oceans \\
\textbf{Ref}: Carbon dioxide also causes ocean acidification because it dissolves in water to form carbonic acid \\

\midrule
\textbf{Claim} Believers think the warming is man-made, while the skeptics believe the warming is natural  and contributions \\from man are minimal and certainly not potentially catastrophic \\
\textbf{\wiki}: The scientific consensus on climate change is that the trend is very likely caused mainly by human-induced emissions \\ of greenhouse gases\\
\textbf{\papers}: Scientists have concluded that the warming observed over the past 50 years is primarily human-induced, and that \\ the effects are ``very likely'' to be catastrophic (although some argue that the effects are likely to be more severe than others) \\
\textbf{Ref}: Attribution sceptics or deniers (who accept the global warming trend but see natural causes for this), \\ and doubt that human activities are responsible for the observed trends.\\

\bottomrule

\end{tabular*}
\caption{Example generated explanations in the \ftof setting based on
  \wiki and \papers (with reference explanations).} 
\label{tab:gen-ex1}
\end{table*}

\subsection{RQ4: What is the optimal number of retrieved documents?}

To check the effect of the number of retrieved documents for both \bm
and \bpr, we present \acc at different retrieval depths $k$ (between 1
and 20) in \figref{res-ret}. We see that for both \fevtwo and \fevthree,
in the case of \bpr we achieve the best performance with 5 documents,
before dropping slightly and flattening out. In the case of \bm, it
takes more retrieved documents (10) to reach the best performance,
before either flattening out or dropping back in performance, suggesting
that the retrieval quality of \bpr is higher than \bm for small values
of $k$.

\begin{table*}[t]
\small
\centering
\begin{tabular*}{\linewidth}{l}\toprule
\textbf{Text} \\
\midrule
\textbf{Claim}: Sea level rise is decelerating. \\
\textbf{\eks} \wiki: Sea levels are rising due to anthropogenic global warming, which is the cause of many global sea level rises.\\ The  current rate of sea level rise is accelerating, and the acceleration of ice loss is due to land-based ice melting.\\
\textbf {Ref}: This acceleration is due mostly to human-caused global warming, which is driving thermal expansion of seawater \\ and the melting of land-based ice sheets and glaciers.\\
\hdashline
\textbf{Claim}: The Great Barrier Reef is in fine fettle \\
\textbf{\eks} \papers: Increased sea surface temperatures boosted by heatwaves have triggered three global mass bleaching \\events in the Great Barrier Reef since 1998.\\
\textbf{Ref}: The percentage of baby corals being born on the Great Barrier Reef dropped drastically in 2018 and scientists are \\ describing it as the early stage of a ``huge natural selection event unfolding''\\

\bottomrule

\end{tabular*}
\caption{Example generated explanations in the \feedtof setting based on
  \wiki and \papers (with reference explanations).}
\label{tab:gen-ex2}
\end{table*}

\begin{table*}[t]
\small
\centering
\begin{tabular*}{\linewidth}{l}\toprule
\textbf{Text} \\
\midrule
 \textbf{Claim}: Marine life has nothing whatsoever to fear from ocean acidification. \textbf{Label}: Refutes \\
\textbf{Trained on \clfev}: Acidification of the oceans has a negative impact on marine ecosystems \textbf{\brscore} 0.46 \\
\textbf{Trained on \clfeed}: Decreasing ocean pH is documented to pose significant risks to marine ecosystems, though the \\ magnitude of the impacts depends on specific species. \textbf{\brscore} 0.33 \\
\textbf {Ref1}: Human activities affect marine life and marine habitats through overfishing, pollution, acidification and the \\ introduction of invasive species.\\
\textbf{Ref2}: Rising levels of acids in seas may endanger marine life, says study \\
\midrule
\textbf{Claim}: Tuvalu sea level isn't rising. \textbf{Label}: Refutes\\
\textbf{Trained on \clfev}:  Tuvalu is affected by the effects of the Perigean spring tide events, which raise the \\ sea level   \textbf{\brscore} 0.75\\
\textbf{Trained on \clfeed} : Global average sea level is rising due to greenhouse gas emissions, with the highest rates in the \\ tropical Pacific, which are vulnerable to coastal erosion \textbf{\brscore} 0.22  \\
\textbf{Ref}: Tuvalu is also affected by perigean spring tide events which raise the sea level higher than a normal high tide.\\
\bottomrule

\end{tabular*}
\caption{Comparative examples in the \ftof and \feedtof settings, with their \brscore.}
\label{tab:gen-ex3}
\end{table*}

\subsection{Generation analysis}

Noting that the generation evaluation metrics (\brscore, \rgone, and
\rgl) may not tell the whole story, we present some example generations
in \tabref{gen-ex1} and \tabref{gen-ex2}. Looking at the first part of
\tabref{gen-ex1}, we see that the generated explanations are similar to the
reference, whereas in second part of the table, the
generated outputs from both \eks (for the same claim) are different to the
reference generation, showing the impact of the two knowledge
sources.

Next we look at the examples in \tabref{gen-ex2}. We see that even
though the model was trained on \clfeed with only 90 training instances,
as a result of the knowledge sources and retrievers, the model can still
generate both coherent and semantically relevant explanations for the
claim, pointing to the fact that high-quality paired data can pay rich
dividends even in small quantities.

\section{Discussion}
 
We present two examples of explanation generation \tabref{gen-ex3} for
\ftof and \feedtof setting, with their \brscore. Looking at the first
example, we see that the \brscore when training on \clfev is higher than
when training on \clfeed. The explanation for \clfev is broadly correct
but contains little detail. On the other hand, the explanation trained
on \clfeed has more substance, as it talks about \textit{pH} and
\textit{species}, but ends up with a lower \brscore. In the second
example, similarly we see a high \brscore for the explanation trained on
\clfev as it is able to extract almost the same evidence as the
reference. This is mainly due to the training and testing set being from
same dataset,\footnote{In this case, the \eks is also \wiki, which was
  used to construct \clf.} allowing the model to potentially extract
exact chunks from context passages. Looking to the explanation from
\clfeed, however, we can see it has at least the same level of
correctness and diversity, but ends up with a lower score.

Given concerns about \brscore being able to account for topical nuances
in the domain of climate change, we additionally performed manual
annotation of the quality of the explanations. Taking inspiration from
\newcite{atanasova+:2020}, we conduct this evaluation in two forms: (1)
given a claim, a generated explanation, and the true veracity label,
annotate whether the explanation is in agreement with the true label, as
a binary classification task (\anone); and (2) rank the two explanations
(for the same claim) according to their overall quality (\antwo).

 \begin{table}[t] \centering
\begin{adjustbox}{max width=0.8\linewidth}
\begin{tabular}{lcc}
\toprule

Dataset                      & \mar & \vag \\
\midrule
\clfev          & 1.64                    & 0.72            \\

\clfeed         & 1.36                     & 0.80           \\

\bottomrule
\end{tabular}
\end{adjustbox}
\caption{Manual evaluation of explanation quality.}
\label{tab:results-ds}
\end{table}

We conduct this manual evaluation on a small sample of 25 examples from
test set, and collect the explanations from the 2 data sources
(resulting if a total of 50 instances across the 2 data sources for the
same claim), shuffling the explanations randomly. We use 3 annotators
and calculate Krippendorff's inter-annotator agreement,\footnote{We use
  nominal metric for \anone and ordinal metric for \antwo}
\cite{hayes+:2007} resulting in $\alpha = 0.58$ for \anone, and
$\alpha = 0.61$ for \antwo. For our evaluation, we then take the
majority vote across annotators. For \anone, we calculate raw agreement
with the true label (\vag; higher is better), and for \antwo we
calculate mean average rank (\mar) across all the 25 examples for each
dataset (lower is better).

Looking at the results in \tabref{results-ds}, we can see that \clfeed
performs better than \clfev over both metrics, in contradiction to what
we found using \brscore, \rgone, and \rgl in \secref{rq3}. This
suggests: (1) that while automatic metrics provide a general sense of
the quality of the generated explanations, they are not able to capture
the subtle nuances of the data; and (2) training the model using
high-quality manually-written explanations, even in small quantities, is
beneficial.

\section{Conclusion}
\label{sec:conclusion}

We explore the task of joint veracity prediction and explanation generation for climate change claims and prepared a data pipeline consisting of knowledge source and paired data. We transposed the idea of using an external knowledge source with with a retriever and a reader from the domain of open domain question answering to for our task and experimented with 2 different knowledge sources, retrievers and number of retrieved documents.  We analysed shortcomings in automatic evaluation like \brscore with the help of manual evaluation and suggested that training model on small high quality manually written explanations augmented with a knowledge source can be quite useful.



\bibliographystyle{acl_natbib}
\bibliography{acl2018,papers}

\end{document}